\documentclass[11pt]{article}

\usepackage[preprint]{acl}

\usepackage{times}
\usepackage{latexsym}

\usepackage[T1]{fontenc}

\usepackage[utf8]{inputenc}

\usepackage{microtype}

\usepackage{inconsolata}

\usepackage{graphicx}
\usepackage{amsmath,amssymb,amsfonts}
\usepackage{makecell}
\usepackage{multirow}
\usepackage{pifont}
\usepackage{arydshln}
\usepackage{tcolorbox}
\tcbuselibrary{breakable}

\definecolor{darkgreen}{rgb}{0.0, 0.5, 0.0}
\definecolor{darkred}{RGB}{164.0, 11.0, 7.0}

\title{GeneralThinker: Domain-General Reasoning through Likelihood-Guided Answer-Conditioned Optimization}

\author{Shengmin Piao \\
  Yonsei University \\
  Seoul, South Korea \\
  \texttt{shengminp@yonsei.ac.kr} \\
  \And
  Sanghyun Park\textsuperscript{\textdagger} \\
  Yonsei University \\
  Seoul, South Korea \\
  \texttt{sanghyun@yonsei.ac.kr} \\}

\begin{document}
\maketitle
\renewcommand{\thefootnote}{}
\footnotetext{\textsuperscript{\textdagger} Corresponding author.}
\renewcommand{\thefootnote}{\arabic{footnote}}
\begin{abstract}
Reinforcement learning with verifiable rewards improves language model reasoning, but its reliance on domain-specific verifiers, sparse outcome rewards, and coarse-grained credit assignment limits its applicability. We introduce \textbf{GeneralThinker}, an on-policy framework that reformulates reasoning supervision as dense answer-conditioned optimization, enabling response-level evaluation and token-level credit assignment without domain-specific verifiers. GeneralThinker evaluates generated reasoning trajectories using the likelihood of the ground-truth answer and derives token-wise compatibility signals for fine-grained credit assignment. To stabilize optimization, it constrains token-level updates through clipping and direction-preserving modulation. Across 11 benchmarks spanning mathematics, STEM, and general reasoning, GeneralThinker achieves the best average performance. Further analyses show that uncontrolled token-level modulation can destabilize training, whereas controlled modulation makes fine-grained credit assignment consistently effective\footnote{Code will be released later.}.
\end{abstract}

\section{Introduction}
Recent progress in language reasoning models has been driven by two complementary paradigms: reinforcement learning for reasoning during training \cite{jaech2024openai} and increased test-time compute during inference \cite{snell2024scaling}. In particular, reinforcement learning with verifiable rewards (RLVR) has become a mainstream training-time approach for improving reasoning capabilities \cite{xu2025toward}, enabling models to acquire advanced behaviors such as dynamic strategy adaptation, self-reflection, and verification \cite{guo2025deepseek}.

\begin{figure}[t]
  \centering
  \includegraphics[width=0.9\columnwidth]{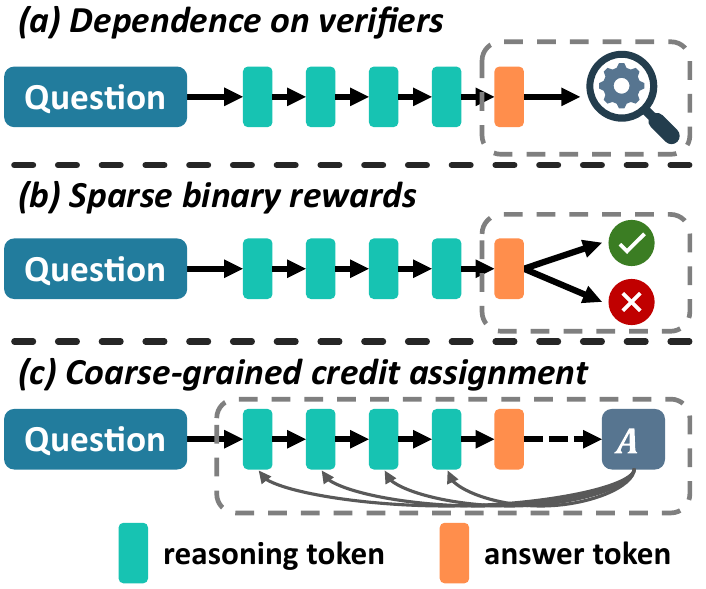}
  \caption{
    Illustration of three key challenges in RLVR-based methods:
    \textbf{(a)} dependence on domain-specific verifiers for reward calculation;
    \textbf{(b)} sparse binary outcome rewards; and
    \textbf{(c)} assigning the same advantage uniformly to all tokens.
    }
  \label{fig:issues}
\end{figure}

Despite these advances, current RLVR-based methods still face three fundamental challenges, as illustrated in Figure~\ref{fig:issues}.

\textbf{(i) Dependence on domain-specific verifiers}: Existing methods are most effective in domains where final answers can be checked automatically and reliably, such as mathematics and coding \cite{zhou2026reinforcing}. However, verification rules designed for these domains do not always transfer across diverse reasoning domains \cite{ma2026generalreasoner}. Thus, broader RLVR training requires reward signals that are less verifier-specific.

\textbf{(ii) Sparse binary outcome rewards}: Most methods optimize response-level rewards based only on final-answer correctness, often in binary form \cite{kwiatkowski2026likelihood}. Because such rewards are assigned only after the full response is generated, responses with substantially different reasoning quality may receive the same reward \cite{yu2025rlpr}. Outcome-only supervision therefore provides only weak discrimination among responses that differ in reasoning quality.

\textbf{(iii) Coarse-grained credit assignment}: Many methods assign a single response-level advantage uniformly to all generated tokens, ignoring quality variation within reasoning trajectories \cite{kazemnejad2025vineppo}. This makes it difficult to identify which parts contribute positively or negatively to the final answer \cite{guo2026segment}. As a result, current RLVR methods often fail to exploit fine-grained supervision signals, leading to information-inefficient optimization for complex responses.

To address these limitations, we propose \textbf{GeneralThinker}, an on-policy framework for training general reasoning models across diverse domains. The central design principle is to derive answer-conditioned training signals from ground-truth answers, providing denser supervision for reasoning trajectories and more localized credit assignment within them.

Specifically, GeneralThinker first uses answer-conditioned likelihood to quantify how strongly a reasoning response supports the ground-truth answer, producing a dense response-level reward. It then uses answer-conditioned token-wise compatibility corrections to modulate the resulting advantage, emphasizing reasoning tokens that are more consistent with the ground-truth answer while down-weighting less compatible ones. Thus, response-level evaluation and token-level advantage modulation jointly provide dense and fine-grained supervision for training. To further stabilize training, GeneralThinker constrains this modulation with token-level clipping and direction-preserving modulation.

To evaluate whether these design choices translate into domain-general reasoning improvements, we conduct controlled experiments under a fixed training setting. All RL-based methods are trained on the same data and are evaluated on 11 benchmarks covering mathematics, STEM, and general reasoning. GeneralThinker achieves the best overall performance across all benchmark domains, with particularly strong gains on challenging benchmarks such as AMC23 and GPQA~\citep{rein2023gpqa}. Further analyses show that the method remains effective across model scales from 1.5B to 7B parameters, and that token-level clipping and direction-preserving modulation are crucial for making token-level advantage modulation stable and consistently beneficial.

The key contributions of this work are as follows:
\begin{itemize}
    \item We propose GeneralThinker, a domain-general on-policy framework that converts final-answer supervision into dense answer-conditioned rewards.

    \item We introduce stabilized token-level advantage modulation for fine-grained credit assignment in reasoning trajectories.

    \item Through extensive experiments, we show that dense answer-conditioned rewards improve general-domain reasoning, while token-level clipping and direction-preserving modulation are essential for stable fine-grained credit assignment.
\end{itemize}

\section{Related Work}
\subsection{RLVR for Diverse-Domain Reasoning}
Recent work has extended RLVR methods beyond strictly verifiable domains along two main lines: constructing general-domain reasoning datasets and designing multi-domain training pipelines.

For dataset construction, prior studies have broadened training corpora to cover more diverse domains and question formats, often by collecting heterogeneous data sources and converting them into forms that support reward computation \cite{ma2026generalreasoner, akter-etal-2026-nemotron}. 
For training pipeline design, recent studies have investigated how RLVR should be organized across heterogeneous domains. Representative approaches include joint multi-domain training, sequential training, and analyses of how domain order affects transfer and overall reasoning performance \cite{wang2025nemotron, yang2026domains, cai2026advancing}. 

However, these efforts primarily expand the data and training scope of RLVR, while leaving its intrinsic limitations in reward sparsity and coarse-grained credit assignment largely unexplored. In contrast, our work derives answer-conditioned signals from the ground-truth answer to support both dense response-level evaluation and fine-grained token-level credit assignment.

\begin{figure*}[t]
  \centering
  \includegraphics[width=1\linewidth]{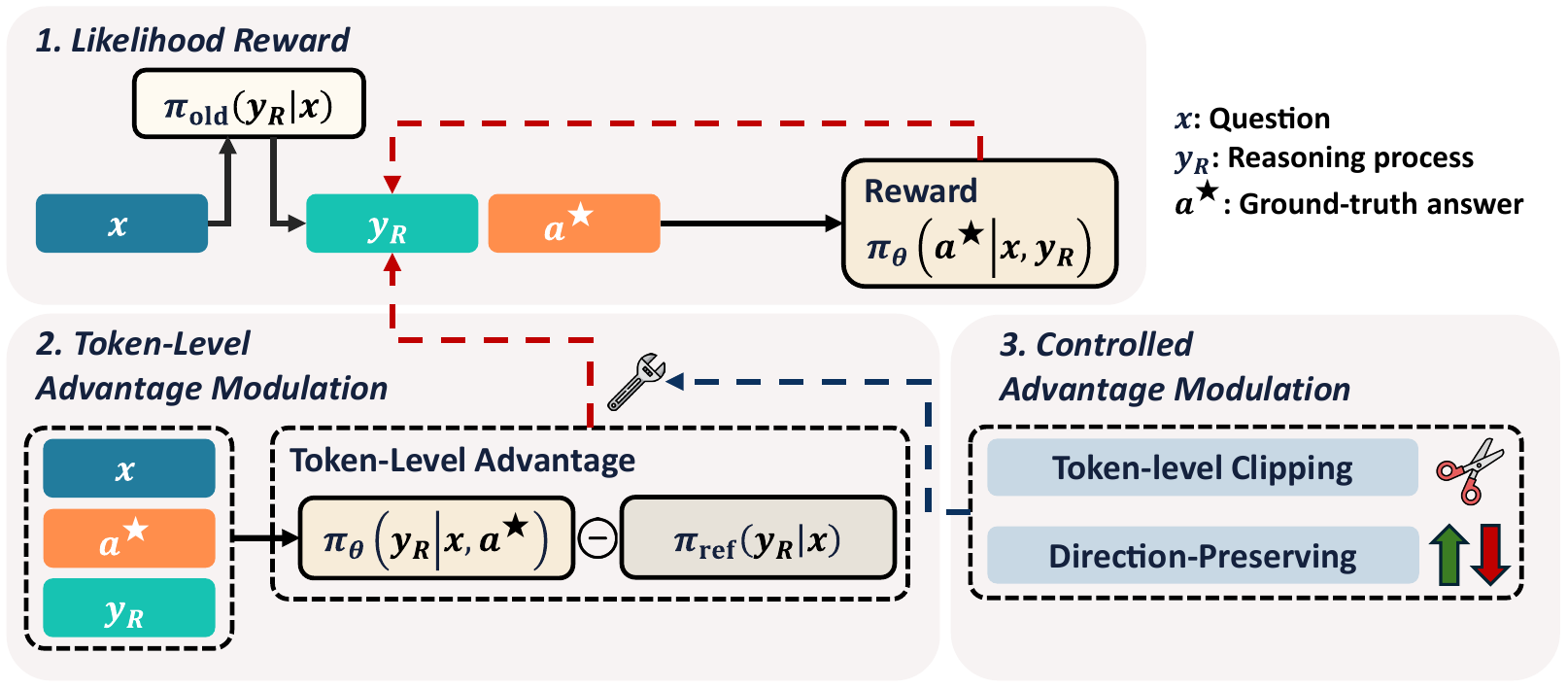}
  \caption{
    Overview of GeneralThinker. The framework first performs response-level evaluation by computing an answer-conditioned likelihood reward, then derives token-wise compatibility signals for token-level advantage modulation, and finally applies token-level clipping and direction-preserving modulation to stabilize fine-grained credit assignment.
    }
  \label{fig:overall framework}
\end{figure*}

\subsection{Answer-Conditioned Reward Design}
Reward design is central to RLVR methods, where supervision is typically provided as outcome-level rewards based on final-answer correctness. While effective in domains such as mathematics and coding, this paradigm relies on reliable verifiers such as symbolic rules, unit tests, or exact matching.

To provide more informative supervision than binary outcome rewards, recent work has explored likelihood-based rewards conditioned on the reference answer. A common strategy is to evaluate a sampled reasoning response by teacher-forcing the ground-truth answer after the generated reasoning, sometimes after explicitly replacing the generated answer with the reference answer \cite{zhou2026reinforcing, yu2025rlpr, liu-etal-2025-nover, kwiatkowski2026likelihood}. These methods reduce reliance on hand-crafted domain-specific verifiers and provide denser response-level rewards than final-answer correctness alone.

Nevertheless, existing answer-conditioned rewards remain primarily response-level signals and therefore offer limited guidance for assigning credit within reasoning trajectories. GeneralThinker extends this line of work by using answer-conditioned token-wise compatibility to modulate response-level advantages, enabling more selective token-level credit assignment.

\subsection{Fine-Grained Credit Assignment}
A central challenge in RLVR methods is credit assignment within generated reasoning trajectories. Ideally, a training method should distinguish productive reasoning tokens or steps from unproductive ones and update the policy accordingly. This issue is particularly important for complex reasoning tasks, where the final answer often depends on multiple intermediate decisions and a single erroneous step can invalidate an otherwise plausible response.

One representative line of work addresses this challenge through process supervision. Process reward models assign scores to intermediate reasoning steps, relying on either human annotations or automated estimation methods to train verifiers for step-level evaluation \cite{lightman2024lets, wang-etal-2024-math}. While these methods provide more informative supervision than outcome-level rewards, they typically require additional annotation, search, or verifier-training procedures.

Compared with process-level supervision, token-level supervision remains less explored in RLVR methods, particularly when the training signal is obtained without training an additional process reward model. GeneralThinker addresses this gap by deriving token-level modulation signals directly from existing policies, without requiring process annotations or additional verifier training, thereby enabling fine-grained token-level credit assignment.

\section{Methodology}

This section presents the framework of GeneralThinker as shown in Figure~\ref{fig:overall framework}. We first describe the fundamental training process (§\ref{subsec:grpo}), followed by the likelihood-based reward (§\ref{subsec:likelihood_reward}). We then detail token-level advantage modulation (§\ref{subsec:advantage_modulation}), and finally introduce controlled advantage modulation (§\ref{subsec:controlled_modulation}).

\subsection{Preliminaries}
Following the notation of GRPO \cite{shao2024deepseekmath}, we distinguish three policies during training. For a given prompt $x$, the policy model $\pi_{\theta}(\cdot \mid x)$ denotes the current trainable model being optimized. The old policy model $\pi_{\text{old}}(\cdot \mid x)$ is used to generate responses, providing samples for the current update. The reference policy $\pi_{\text{ref}}(\cdot \mid x)$ is a frozen copy of the initial policy model and is used to regularize optimization.

For the $j$-th sampled response, we write
\[
y^{(j)} = \bigl(y^{(j)}_1, \dots, y^{(j)}_{T_j}\bigr) = \bigl(y^{(j)}_{R}, y^{(j)}_{A}\big)
\]
where $y^{(j)}_{R}$ is the sampled reasoning prefix, $y^{(j)}_{A}$ is the predicted answer, $T_j=|y^{(j)}|$ and $T^{R}_j=|y^{(j)}_{R}|$ denote the lengths of the full response and reasoning prefix, respectively.

\subsection{GRPO Training Process}
\label{subsec:grpo}

For each prompt $x$, we sample a group of $G$ responses from the old policy:
\[
y^{(j)} \sim \pi_{\text{old}}(\cdot \mid x), \qquad j = 1, \dots, G
\]
Each response is evaluated by a verifier $R$, which assigns a reward:
\[
r_j = R(x, y^{(j)},a^\star)
\]
where $a^\star$ denotes the ground-truth answer.

We then compute the group-level mean and standard deviation of the rewards:
\begin{equation}
\label{eq:grpo-mean_std}
\mu_r = \frac{1}{G} \sum_{j=1}^G r_j,
\qquad
\sigma_r = \sqrt{\frac{1}{G} \sum_{j=1}^G (r_j - \mu_r)^2}
\end{equation}
Using these statistics, the group-normalized advantage for response $j$ is defined as
\begin{equation}
\label{eq:grpo-adv}
\widehat{A}_j = \frac{r_j - \mu_r}{\sigma_r}
\end{equation}
This normalized advantage measures the relative performance of $y^{(j)}$ within the sampled group. In practice, it is broadcast to all tokens in the response, which is $\widehat{A}_{j,t} \equiv \widehat{A}_j$ for all $t$.

To update the policy using samples generated by the old policy, each token is reweighted by the likelihood ratio between the current and old policies:
\begin{equation}
\rho_{j,t}(\theta) =
\frac{\pi_{\theta}(y_t^{(j)} \mid x, y_{<t}^{(j)})}
{\pi_{\text{old}}(y_t^{(j)} \mid x, y_{<t}^{(j)})}
\end{equation}
To stabilize training, the clipped surrogate objective from PPO \cite{schulman2017proximal} is applied, which clips the ratio to the interval $[1-\epsilon, 1+\epsilon]$:
\[
\mathrm{clip}(\rho; 1 - \epsilon, 1 + \epsilon) = \min\{1 + \epsilon, \max\{1 - \epsilon, \rho\}\},
\]
\[
\mathrm{CLIP}(\rho, \widehat{A}, \epsilon) =
\min\big(
\rho \widehat{A},\;
\mathrm{clip}(\rho; 1 - \epsilon, 1 + \epsilon)\widehat{A}
\big)
\]
where $\epsilon$ controls the maximum deviation from the old policy in each update.

The per-response objective is defined as
\begin{equation}
S_j(\theta) =\frac{1}{T_j}\sum_{t=1}^{T_j}
\mathrm{CLIP}\left(\rho_{j,t}(\theta), \widehat{A}_{j,t}, \epsilon \right)
\end{equation}
and the overall training objective is then given by
\begin{equation}
\begin{aligned}
\mathcal{J}(\theta)
&=\mathbb{E}_{x, y^{(1:G)}} \left[\frac{1}{G} \sum_{j=1}^G S_j(\theta)\right]
\\
&\quad
- \beta\,\mathbb{E}_{x}
\Big[D_{\mathrm{KL}}\left(\pi_{\theta}(\cdot \mid x)\| \pi_{\mathrm{ref}}(\cdot \mid x)\right)\Big]
\end{aligned}
\end{equation}
where $\beta$ is the regularization weight. Training proceeds by minimizing the negative objective:
\[
L(\theta) = -\mathcal{J}(\theta)
\]

\subsection{Likelihood Reward}
\label{subsec:likelihood_reward}
As discussed earlier, the reward $r_j$ produced by the verifier $R$ is typically binary, which leads to sparse supervision and limited granularity.

To address this limitation, we introduce a likelihood-based reward. Specifically, for each sampled response, we extract the portion preceding \texttt{\textbackslash boxed\{\}} as the reasoning prefix $y^{(j)}_{R}$. We then replace the predicted answer $y^{(j)}_{A}$ inside \texttt{\textbackslash boxed\{\}} with the ground-truth answer $a^\star$, and feed the modified sequence back into the current policy. The likelihood reward is defined as the log-likelihood of generating $a^\star$ conditioned on the prompt $x$ and the reasoning prefix $y^{(j)}_{R}$:

\begin{equation}
\begin{aligned}
r_j^{\mathrm{LR}}
&= \log \pi_{\theta}(a^\star \mid x, y^{(j)}_{R}) \\
&= \sum_{t=1}^{|a^\star|} \log \pi_{\theta}(a_t^\star \mid x, y^{(j)}_{R}, a^\star_{<t})
\end{aligned}
\end{equation}

Intuitively, the likelihood reward quantifies how well a generated reasoning prefix supports the ground-truth answer under the current policy. For a fixed input and answer, a higher log-likelihood indicates that the model assigns greater probability to $a^\star$ after conditioning on $y^{(j)}_{R}$, suggesting that the prefix is more compatible with producing the correct answer under the current policy. In contrast, a lower log-likelihood indicates that the prefix provides limited or misleading guidance. Therefore, the likelihood reward provides a dense, answer-conditioned signal at the response level.

\subsection{Token-Level Advantage Modulation}
\label{subsec:advantage_modulation}

Although the likelihood reward provides dense response-level supervision, the normalized advantage $\widehat{A}^{\mathrm{LR}}_j$ is still shared by all tokens.

To obtain token-level guidance, we construct an answer-conditioned token-wise signal over the reasoning prefix. Let $x^+$ denote an answer-conditioned version of the prompt, where the ground-truth answer $a^\star$ is provided as part of the input\footnote{The detailed prompt is provided in Appendix.}. For each reasoning token, we define
\begin{equation}
\begin{aligned}
r^{\mathrm{tok}}_{j,t}
&=\log \mathrm{sg}\bigl[\pi_{\theta}(y_t^{(j)} \mid x^+, y_{<t}^{(j)})\bigr] 
\\
& -\log \pi_{\mathrm{ref}}(y_t^{(j)} \mid x, y_{<t}^{(j)}), \qquad t = 1,\dots,T_j^{\mathrm{R}}
\end{aligned}
\end{equation}
where $\mathrm{sg}[\cdot]$ denotes the stop-gradient operator. This signal compares the likelihood of each reasoning token under an answer-conditioned context with its likelihood under the reference policy, thereby estimating its relative compatibility with the ground-truth answer.

To use this signal for advantage modulation rather than as an additional response-level advantage, we normalize the token-wise signal within each reasoning prefix. Specifically, we compute
\begin{equation}
\label{eq:tok-mean-std}
\begin{aligned}
\mu^{\mathrm{tok}}_j
&=\frac{1}{T_j^{\mathrm{R}}}\sum_{t=1}^{T_j^{\mathrm{R}}} r^{\mathrm{tok}}_{j,t},
\\
\sigma^{\mathrm{tok}}_j
&=
\sqrt{
\frac{1}{T_j^{\mathrm{R}}} \sum_{t=1}^{T_j^{\mathrm{R}}}
\left(r^{\mathrm{tok}}_{j,t} - \mu^{\mathrm{tok}}_j \right)^2
}
\end{aligned}
\end{equation}
Using these statistics, we define the normalized token-level modulation term as
\begin{equation}
\label{eq:tok-modulation}
\widetilde{\delta}_{j,t} =
\frac{
r^{\mathrm{tok}}_{j,t} - \mu^{\mathrm{tok}}_j}
{\sigma^{\mathrm{tok}}_j},
\qquad
t=1,\dots,T_j^{\mathrm{R}}
\end{equation}
The zero-mean normalization makes $\widetilde{\delta}_{j,t}$ a relative token-level correction, so it redistributes credit within the prefix without changing its average advantage.

Finally, we define the advantage for reasoning tokens in the policy objective as
\begin{equation}
\label{eq:final-token-advantage}
\widehat{A}_{j,t} = \widehat{A}_j^{\mathrm{LR}} + \lambda_{\mathrm{tok}} \widetilde{\delta}_{j,t},
\qquad t=1,\dots,T_j^{\mathrm{R}}
\end{equation}
where $\lambda_{\mathrm{tok}}$ controls the strength of token-level advantage modulation\footnote{Following standard GRPO-style optimization, all rewards, reward-derived advantages, and token-level modulation terms are detached before policy optimization; gradients are propagated only through the current-policy likelihood ratios and the KL regularization term.}.

\subsection{Controlled Advantage Modulation}
\label{subsec:controlled_modulation}

\paragraph{Token-Level Clipping}
In practice, the normalized modulation term $\widetilde{\delta}_{j,t}$ may contain occasional large-magnitude outliers. To prevent a small number of extreme reasoning tokens from dominating training, we apply token-level clipping:
\begin{equation}
\delta^{\mathrm{clip}}_{j,t}
=
\operatorname{clip}\left(
\widetilde{\delta}_{j,t},
-\epsilon_{\mathrm{tok}},
\epsilon_{\mathrm{tok}}
\right)
\end{equation}
where $\epsilon_{\mathrm{tok}}>0$ denotes the clipping threshold.

We then re-center the clipped values within each prefix to preserve the zero-mean property, ensuring that clipping only affects relative token-level credit.

\paragraph{Direction-Preserving Modulation}
An unrestricted additive token-level modulation term may reverse the optimization direction induced by the likelihood advantage $\widehat{A}_j^{\mathrm{LR}}$. To avoid this issue, we transform the token-level modulation signal into a bounded weight. Specifically, we define
\begin{equation}
s_j = \max\left(1, \max_{1\le t\le T_j^{\mathrm{R}}} |\delta^{\mathrm{clip}}_{j,t}|\right),
\end{equation}
and
\begin{equation}
\omega_{j,t}
=
\frac{\delta^{\mathrm{clip}}_{j,t}}{s_j}
\end{equation}

For each reasoning token, the final advantage is then given by
\begin{equation}
\widehat{A}_{j,t}
=
\widehat{A}_j^{\mathrm{LR}}
+
\lambda_{\mathrm{tok}}
|\widehat{A}_j^{\mathrm{LR}}|
\omega_{j,t},
\end{equation}
where $\lambda_{\mathrm{tok}} \in [0,1)$ controls the strength of token-level modulation.

\begin{table*}[t]
\renewcommand{\arraystretch}{0.96}
    \centering
    \begin{tabular}{lcccc}
        \Xhline{1.0pt}
        & \textbf{Base} & \textbf{Binary-RL} & \textbf{Likelihood-RL} & \textbf{GeneralThinker}\\
        \Xhline{0.5pt}
        \multicolumn{5}{c}{\textbf{Math \& STEM}} \\
        MATH-500        & 48.20             & 53.80             & \textbf{54.80}             & \underline{54.00} \\
        GSM8K           & 71.63             & 72.10             & \underline{72.33}          & \textbf{72.40}\\ 
        Minerva         & 15.44             & \underline{20.22} & 19.85                      & \textbf{20.59} \\
        OlympiadBench    & 19.44             & \underline{20.33} & 19.88                      & \textbf{21.07} \\
        AMC23           & 17.50             & \underline{22.50} & 20.00                      & \textbf{30.00} \\
        GPQA            & 14.96             & \underline{16.74} & 15.40                      & \textbf{20.54} \\
        GPQA-Diamond    & 14.14             & 14.65             & \underline{18.18}          & \textbf{19.19} \\
        \cdashline{1-5}
        \textit{Average}    & 28.76         & 31.48             & \underline{31.49}             & \textbf{33.97} \\
        \Xhline{0.5pt}
        \multicolumn{5}{c}{\textbf{General}} \\
        MMLU            & 44.82             & 49.74             & \underline{52.61}          & \textbf{53.50} \\
        MMLU-Pro        & 22.39             & 25.08             & \underline{26.32}          & \textbf{27.13}\\
        MMLU-Redux      & 45.35             & 51.02             & \underline{53.77}          & \textbf{54.72}\\
        SuperGPQA       & 12.26             & 12.40             & \underline{12.76}          & \textbf{12.79} \\
        \cdashline{1-5}
        \textit{Average}    & 31.21         & 34.56             & \underline{36.37}             & \textbf{37.04} \\
        \Xhline{1.0pt}
    \end{tabular}
    \caption{Accuracy (\%) of GeneralThinker compared with baselines. \textit{Average} denotes the unweighted mean across benchmarks within each category. \textbf{Bold} numbers indicate the best performance, and \underline{underlined} numbers indicate the second-best.}
    \label{tab:overall-performance}
\end{table*}

This formulation keeps the response-level likelihood advantage as the primary learning signal while using bounded token-level weights to redistribute credit among reasoning tokens. Since $\omega_{j,t}\in[-1,1]$ and $\lambda_{\mathrm{tok}}<1$, the token-level term cannot reverse the sign of $\widehat{A}_j^{\mathrm{LR}}$.

\section{Experimental Setup}
\subsection{Datasets and Metrics}
To expose all RL-based methods to diverse domains and question formats during training, we construct a cross-domain training set from two complementary sources: NEMOTRON-CROSSTHINK-QA~\citep{akter-etal-2026-nemotron}, a general-domain reasoning dataset, and MATH~\citep{hendrycks2021measuring}, a mathematical reasoning dataset. We use the same training data for all baselines, so that performance differences primarily reflect the training objective rather than data composition. Detailed dataset information is provided in Appendix~\ref{sec:datasets}.

We evaluate performance using answer-level accuracy, defined as the proportion of instances for which the final predicted answer exactly matches the corresponding ground-truth answer across all benchmarks. Unless otherwise specified, all responses are generated using greedy decoding.

\subsection{Benchmarks and Baselines}
To evaluate whether GeneralThinker is effective across diverse domains rather than being confined to narrowly verifiable settings, we construct an evaluation suite spanning general-domain, mathematical, and STEM reasoning benchmarks. The general-domain benchmarks assess broad general reasoning capabilities, while the mathematical and STEM benchmarks examine its performance on widely used quantitative and scientific reasoning tasks. We compare GeneralThinker with several controlled baselines to evaluate its effectiveness under consistent training and evaluation settings. Detailed descriptions of the benchmarks and baselines are provided in Appendix~\ref{sec:benchmarks} and Appendix~\ref{sec:baselines}, respectively.

\subsection{Implementation Details}
We use Qwen2.5-1.5B/3B/7B~\citep{Yang2024Qwen25TR} as backbone models and apply LoRA fine-tuning~\citep{hu2022lora} to train GeneralThinker\footnote{We do not use higher-capacity models such as Qwen3 because they are typically specialized for reasoning. To better isolate the effect of our fine-tuning method, we instead use the aforementioned models.}. Training is conducted on four A100 GPUs using the Hugging Face Transformers library~\citep{wolf2020transformers}, initialized from publicly released pretrained checkpoints\footnote{\url{https://huggingface.co/collections/Qwen/qwen25}}. During training, we use vLLM~\citep{kwon2023efficient} for efficient response generation. Complete hyperparameter settings are provided in Appendix~\ref{sec:hyperparameter}.

\section{Results and Analysis}
\subsection{Overall Performance}
\label{subsec:overall_performance}
Table~\ref{tab:overall-performance} presents the overall evaluation results of GeneralThinker against the base model and two RL-based baselines. We report the accuracy on each benchmark, together with the unweighted average within each benchmark group, to provide both fine-grained comparisons and an overall summary of performance trends.

Overall, all RL-based methods improve upon the base model across the evaluated benchmarks. In the Math \& STEM domain, Binary-RL and Likelihood-RL improve the average accuracy from 28.76\% to 31.48\% and 31.49\%, respectively. In the General domain, the average accuracy increases from 31.21\% to 34.56\% for Binary-RL and 36.37\% for Likelihood-RL. These results show that RL-based optimization consistently enhances the reasoning ability of the base model across all tasks.

Comparing the two RL baselines, Binary-RL and Likelihood-RL exhibit different strengths across domains. On the Math \& STEM benchmarks, the two methods achieve nearly identical average scores, with each exhibiting advantages on different benchmarks. This suggests that neither reward design consistently dominates in mathematical and STEM reasoning. By contrast, Likelihood-RL consistently outperforms Binary-RL on all general reasoning benchmarks, which suggests that likelihood-based rewards provide a more useful response-level signal than binary rewards for the evaluated general-domain reasoning tasks.

\begin{figure}[t]
  \centering
  \includegraphics[width=0.9\linewidth]{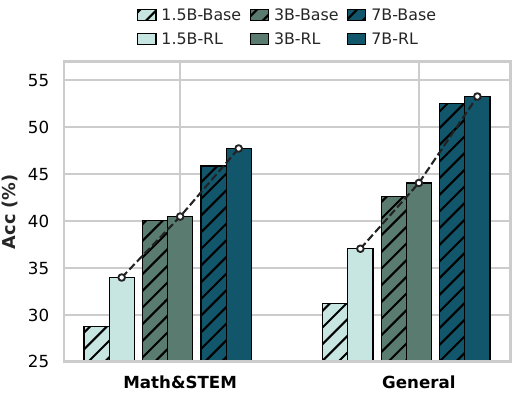}
  \caption{Scalability of GeneralThinker across model sizes in Math \& STEM and General domains. Dashed lines connect RL-enhanced results across scales.}
  \label{fig:scalability}
\end{figure}

GeneralThinker achieves the strongest overall performance across both benchmark groups. In Math \& STEM, it reaches an average accuracy of 33.97\%, outperforming the strongest RL baseline by 2.48\% points. The gains are particularly notable on challenging benchmarks such as AMC23 and GPQA. In the General domain, GeneralThinker further improves the average accuracy to 37.04\%, surpassing Binary-RL and Likelihood-RL by 2.48\% and 0.67\% points, respectively.

Taken together, these results demonstrate that GeneralThinker provides the most robust overall improvement. Compared with standard RL baselines, our method achieves higher average performance in both Math \& STEM and General domains, while also obtaining the best results on most individual benchmarks. This indicates that GeneralThinker improves reasoning performance not only on specialized mathematical and STEM tasks, but also on broader general reasoning tasks.

\begin{table}[!t]
\renewcommand{\arraystretch}{0.95}
    \centering
    \resizebox{\linewidth}{!}{
        \begin{tabular}{ccccc}
        \Xhline{1.0pt}
        \textbf{AM} & \textbf{TC} & \textbf{DP} & \textbf{Math \& STEM} & \textbf{General}\\
        \Xhline{0.5pt}
        \textcolor{red}{\ding{55}}       & \textcolor{red}{\ding{55}}       & \textcolor{red}{\ding{55}}       & 31.49     & 36.37 \\
        \textcolor{darkgreen}{\ding{51}} & \textcolor{red}{\ding{55}}       & \textcolor{red}{\ding{55}}       & 31.93     & 34.31 \\
        \textcolor{darkgreen}{\ding{51}} & \textcolor{darkgreen}{\ding{51}} & \textcolor{red}{\ding{55}}       & 32.98     & 34.53 \\
        \textcolor{darkgreen}{\ding{51}} & \textcolor{red}{\ding{55}}       & \textcolor{darkgreen}{\ding{51}} & 32.28     & 35.07 \\
        \textcolor{darkgreen}{\ding{51}} & \textcolor{darkgreen}{\ding{51}} & \textcolor{darkgreen}{\ding{51}} & \textbf{33.97}     & \textbf{37.04} \\
        \Xhline{1.0pt}
        \end{tabular}
    }
    \caption{Ablation results of GeneralThinker; scores are average accuracy (\%) within each benchmark domain.}
    \label{tab:ablation_study}
\end{table}

\begin{figure*}[t]
  \centering
  \includegraphics[width=1\linewidth]{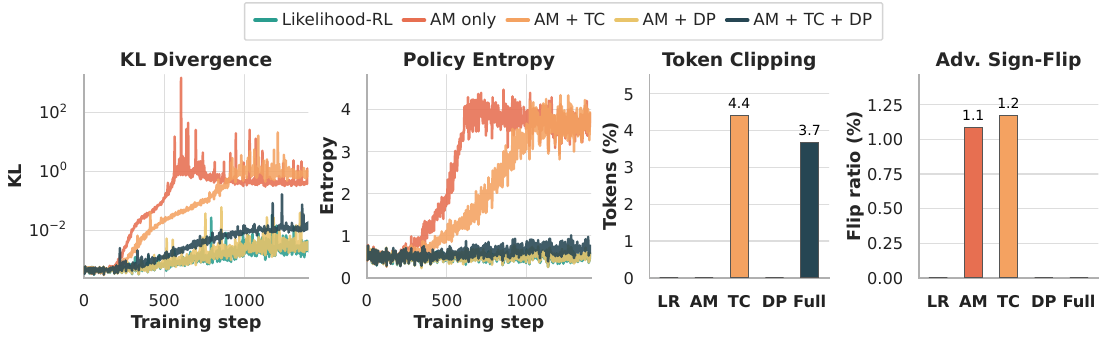}
  \caption{Training stability analysis under different stabilization settings.}
  \label{fig:analysis}
\end{figure*}

\paragraph{Method Scalability} We further evaluate whether the effectiveness of GeneralThinker is preserved across model scales. As shown in Figure~\ref{fig:scalability}, GeneralThinker consistently improves over the corresponding base model at all evaluated scales in both the Math \& STEM and General domains. In addition, the average accuracy of the RL-enhanced models increases steadily as the model size grows, indicating that GeneralThinker remains effective when applied to larger models. These results suggest that GeneralThinker shows consistent scalability across model sizes. Detailed benchmark results are provided in Appendix~\ref{sec:datailed_results}.

\subsection{Ablation Study}
\label{subsec:ablation_study}
We conduct a series of ablation experiments to disentangle the effects of GeneralThinker's three core components: token-level advantage modulation (AM), token-level clipping (TC), and direction-preserving modulation (DP). The baseline is trained only with the likelihood reward. We then add token-level advantage modulation to examine whether fine-grained credit assignment is effective without stabilization. Based on this variant, token-level clipping and direction-preserving modulation are introduced separately to assess the effect of each stabilization mechanism. Finally, all components are combined to evaluate their joint contribution.

Table~\ref{tab:ablation_study} shows that applying token-level advantage modulation alone does not lead to consistent gains. While it slightly improves performance in the Math \& STEM domain, it degrades performance in the General domain. This suggests that fine-grained token-level signals are not sufficient by themselves: without proper stabilization, they may introduce noisy or overly aggressive updates.

Introducing stabilization mechanisms alleviates this issue. Both token-level clipping and direction-preserving modulation improve the variant with token-level advantage modulation, particularly in the Math \& STEM domain, and partially recover the General domain performance. This indicates that controlling the magnitude and direction of token-level modulation is important for making fine-grained credit assignment effective.

The full model achieves the highest average accuracy in both domains. This demonstrates that the three components are complementary: token-level advantage modulation provides fine-grained credit assignment, while token-level clipping and direction-preserving modulation stabilize its optimization effect. Overall, the ablation results confirm that GeneralThinker benefits from combining token-level modulation with controlled stabilization. Detailed benchmark results are provided in Appendix~\ref{sec:datailed_results}.

\subsection{Analysis Study}
\label{subsec:analysis_study}

Figure~\ref{fig:analysis} compares the training dynamics under different stabilization settings. Using token-level advantage modulation alone leads to unstable optimization. The KL divergence exhibits large fluctuations, suggesting that the model distribution is repeatedly pushed away from the reference policy during training. Meanwhile, the policy entropy also fluctuates sharply, indicating unstable exploration and inconsistent policy uncertainty. These two trends suggest that unconstrained token-level modulation not only increases distributional drift, but also disrupts the policy's exploration behavior.

This instability is substantially reduced when either token-level clipping or direction-preserving modulation is introduced. The two mechanisms mitigate instability in complementary ways. Token-level clipping mainly reduces severe KL spikes, although some oscillations remain. Direction-preserving modulation provides a stronger stabilizing effect, leading to a smoother KL response. Combining the two yields the most stable KL curve, with only mild oscillations in later training stages.

The entropy trajectories show a consistent pattern. Token-level clipping reduces extreme entropy fluctuations, although clear oscillations remain. Direction-preserving modulation provides a stronger stabilizing effect, leading to a smoother entropy response. Combining the two yields the most consistent entropy curve, suggesting that controlled token-level modulation supports fine-grained credit assignment without inducing erratic exploration.

A further observation is that the interventions are sparse. The average token clipping ratio is below 5\%, and the average sign-flip ratio is approximately 1\%. This indicates that only a small fraction of tokens trigger these constraints. Nevertheless, these rare exceptional updates are sufficient to gradually undermine training stability if left uncontrolled. Therefore, token-level clipping and direction-preserving modulation do not act as aggressive regularizers over all tokens; instead, they selectively suppress high-risk updates that can accumulate and destabilize token-level advantage modulation.

\section{Conclusion}
We presented GeneralThinker, a domain-general on-policy framework for improving language model reasoning with dense answer-conditioned supervision and stabilized token-level advantage modulation. GeneralThinker consistently improves performance across mathematics, STEM, and general reasoning benchmarks, and remains effective across model scales. Our analysis further shows that controlled token-level modulation is important for stable fine-grained credit assignment. These results highlight dense answer-conditioned rewards as a promising direction for extending RL-based reasoning training beyond narrowly verifiable domains.

\section{Limitations}
GeneralThinker derives token-level modulation signals from answer-conditioned likelihood differences, which provide an intuitive proxy for token-answer compatibility. However, these signals may not guarantee the true causal contribution of each token to final-answer correctness. Although subtracting the reference-policy likelihood and applying zero-mean normalization reduce general fluency biases and preserve the response-level advantage on average, the signal may still correlate with spurious lexical, formatting, or answer-revealing cues.

This limitation suggests a direction for future work. Not all reasoning tokens are equally informative for credit assignment: many tokens are predictable syntactic or formatting continuations and may contribute little to reasoning quality. Recent studies suggest that useful learning signals are often concentrated in high-entropy tokens, where the policy is more uncertain and alternative choices may induce different reasoning paths \citep{wang2026beyond}. Future work could therefore explore entropy-gated modulation, applying fine-grained token-level updates only to uncertain and potentially decision-critical tokens while preserving the response-level advantage for other tokens.

\bibliography{custom}

@article{jaech2024openai,
    title={Openai o1 system card},
    author={Jaech, Aaron and Kalai, Adam and Lerer, Adam and Richardson, Adam and El-Kishky, Ahmed and Low, Aiden and Helyar, Alec and Madry, Aleksander and Beutel, Alex and Carney, Alex and others},
    journal={arXiv preprint arXiv:2412.16720},
    year={2024},
    url={https://arxiv.org/abs/2412.16720}
}

@inproceedings{snell2024scaling,
    title={Scaling {LLM} Test-Time Compute Optimally Can be More Effective than Scaling Parameters for Reasoning},
    author={Charlie Victor Snell and Jaehoon Lee and Kelvin Xu and Aviral Kumar},
    booktitle={The Thirteenth International Conference on Learning Representations},
    year={2025},
    url={https://openreview.net/forum?id=4FWAwZtd2n}
}

@article{guo2025deepseek,
    title={Deepseek-r1: Incentivizing reasoning capability in llms via reinforcement learning},
    author={Guo, Daya and Yang, Dejian and Zhang, Haowei and Song, Junxiao and Zhang, Ruoyu and Xu, Runxin and Zhu, Qihao and Ma, Shirong and Wang, Peiyi and Bi, Xiao and others},
    journal={arXiv preprint arXiv:2501.12948},
    year={2025},
    url={https://arxiv.org/abs/2501.12948}
}

@article{xu2025toward,
    title={Toward large reasoning models: A survey of reinforced reasoning with large language models},
    author={Xu, Fengli and Hao, Qianyue and Shao, Chenyang and Zong, Zefang and Li, Yu and Wang, Jingwei and Zhang, Yunke and Wang, Jingyi and Lan, Xiaochong and Gong, Jiahui and others},
    journal={Patterns},
    volume={6},
    number={10},
    year={2025},
    publisher={Elsevier}
}

@article{kwiatkowski2026likelihood,
    title={Likelihood-Based Reward Designs for General LLM Reasoning},
    author={Kwiatkowski, Ariel and Butt, Natasha and Labiad, Ismail and Kempe, Julia and Ollivier, Yann},
    journal={arXiv preprint arXiv:2602.03979},
    year={2026}
}

@article{yu2025rlpr,
    title={Rlpr: Extrapolating rlvr to general domains without verifiers},
    author={Yu, Tianyu and Ji, Bo and Wang, Shouli and Yao, Shu and Wang, Zefan and Cui, Ganqu and Yuan, Lifan and Ding, Ning and Yao, Yuan and Liu, Zhiyuan and others},
    journal={arXiv preprint arXiv:2506.18254},
    year={2025}
}

@inproceedings{ma2026generalreasoner,
    title={General-Reasoner: Advancing {LLM} Reasoning Across All Domains},
    author={Xueguang Ma and Qian Liu and Dongfu Jiang and Ge Zhang and Zejun MA and Wenhu Chen},
    booktitle={The Thirty-ninth Annual Conference on Neural Information Processing Systems},
    year={2026},
    url={https://openreview.net/forum?id=pBFVoll8Xa}
}

@inproceedings{zhou2026reinforcing,
    title={Reinforcing General Reasoning Without Verifiers},
    author={Xiangxin Zhou and Zichen Liu and Anya Sims and Haonan Wang and Tianyu Pang and Chongxuan Li and Liang Wang and Min Lin and Chao Du},
    booktitle={The Fourteenth International Conference on Learning Representations},
    year={2026},
    url={https://openreview.net/forum?id=nnwvwge40d}
}

@inproceedings{kazemnejad2025vineppo,
    title={Vine{PPO}: Refining Credit Assignment in {RL} Training of {LLM}s},
    author={Amirhossein Kazemnejad and Milad Aghajohari and Eva Portelance and Alessandro Sordoni and Siva Reddy and Aaron Courville and Nicolas Le Roux},
    booktitle={Forty-second International Conference on Machine Learning},
    year={2025},
    url={https://openreview.net/forum?id=Myx2kJFzAn}
}

@inproceedings{guo2026segment,
    title={Segment Policy Optimization: Effective Segment-Level Credit Assignment in {RL} for Large Language Models},
    author={Yiran Guo and Lijie Xu and Jie Liu and Ye Dan and Shuang Qiu},
    booktitle={The Thirty-ninth Annual Conference on Neural Information Processing Systems},
    year={2026},
    url={https://openreview.net/forum?id=9osvTOYbT4}
}

@inproceedings{akter-etal-2026-nemotron,
    title = "Nemotron-{C}ross{T}hink: Scaling Self-Learning beyond Math Reasoning",
    author = "Akter, Syeda Nahida  and
      Prabhumoye, Shrimai  and
      Novikov, Matvei  and
      Han, Seungju  and
      Lin, Ying  and
      Bakhturina, Evelina  and
      Nyberg, Eric  and
      Choi, Yejin  and
      Patwary, Mostofa  and
      Shoeybi, Mohammad  and
      Catanzaro, Bryan",
    editor = "Demberg, Vera  and
      Inui, Kentaro  and
      Marquez, Llu{\'i}s",
    booktitle = "Proceedings of the 19th Conference of the {E}uropean Chapter of the {A}ssociation for {C}omputational {L}inguistics (Volume 1: Long Papers)",
    month = mar,
    year = "2026",
    address = "Rabat, Morocco",
    publisher = "Association for Computational Linguistics",
    url = "https://aclanthology.org/2026.eacl-long.43/",
    doi = "10.18653/v1/2026.eacl-long.43",
    pages = "984--1002",
    ISBN = "979-8-89176-380-7",
}

@article{wang2025nemotron,
  title={Nemotron-cascade: Scaling cascaded reinforcement learning for general-purpose reasoning models},
  author={Wang, Boxin and Lee, Chankyu and Lee, Nayeon and Lin, Sheng-Chieh and Dai, Wenliang and Chen, Yang and Chen, Yangyi and Yang, Zhuolin and Liu, Zihan and Shoeybi, Mohammad and others},
  journal={arXiv preprint arXiv:2512.13607},
  year={2025}
}

@article{yang2026domains,
  title={When Domains Interact: Asymmetric and Order-Sensitive Cross-Domain Effects in Reinforcement Learning for Reasoning},
  author={Yang, Wang and Wang, Shouren and Song, Chaoda and Ma, Chuang and Li, Xinpeng and Wang, Nengbo and Zhou, Kaixiong and Chaudhary, Vipin and Han, Xiaotian},
  journal={arXiv preprint arXiv:2602.01365},
  year={2026}
}

@article{cai2026advancing,
  title={Advancing General-Purpose Reasoning Models with Modular Gradient Surgery},
  author={Cai, Min and Liang, Yu and Wang, Longzheng and Wang, Yan and Zhang, Yueyang and Xia, Long and Sun, Zhiyuan and Ye, Xi and Shi, Daiting},
  journal={arXiv preprint arXiv:2602.02301},
  year={2026}
}

@inproceedings{liu-etal-2025-nover,
    title = "{NOVER}: Incentive Training for Language Models via Verifier-Free Reinforcement Learning",
    author = "Liu, Wei  and
      Qi, Siya  and
      Wang, Xinyu  and
      Qian, Chen  and
      Du, Yali  and
      He, Yulan",
    editor = "Christodoulopoulos, Christos  and
      Chakraborty, Tanmoy  and
      Rose, Carolyn  and
      Peng, Violet",
    booktitle = "Proceedings of the 2025 Conference on Empirical Methods in Natural Language Processing",
    month = nov,
    year = "2025",
    address = "Suzhou, China",
    publisher = "Association for Computational Linguistics",
    url = "https://aclanthology.org/2025.emnlp-main.378/",
    doi = "10.18653/v1/2025.emnlp-main.378",
    pages = "7439--7458",
    ISBN = "979-8-89176-332-6",
}

@inproceedings{lightman2024lets,
    title={Let's Verify Step by Step},
    author={Hunter Lightman and Vineet Kosaraju and Yuri Burda and Harrison Edwards and Bowen Baker and Teddy Lee and Jan Leike and John Schulman and Ilya Sutskever and Karl Cobbe},
    booktitle={The Twelfth International Conference on Learning Representations},
    year={2024},
    url={https://openreview.net/forum?id=v8L0pN6EOi}
}

@inproceedings{wang-etal-2024-math,
    title = "Math-Shepherd: Verify and Reinforce {LLM}s Step-by-step without Human Annotations",
    author = "Wang, Peiyi  and
      Li, Lei  and
      Shao, Zhihong  and
      Xu, Runxin  and
      Dai, Damai  and
      Li, Yifei  and
      Chen, Deli  and
      Wu, Yu  and
      Sui, Zhifang",
    editor = "Ku, Lun-Wei  and
      Martins, Andre  and
      Srikumar, Vivek",
    booktitle = "Proceedings of the 62nd Annual Meeting of the Association for Computational Linguistics (Volume 1: Long Papers)",
    month = aug,
    year = "2024",
    address = "Bangkok, Thailand",
    publisher = "Association for Computational Linguistics",
    url = "https://aclanthology.org/2024.acl-long.510/",
    doi = "10.18653/v1/2024.acl-long.510",
    pages = "9426--9439",
}

@article{shao2024deepseekmath,
    title={Deepseekmath: Pushing the limits of mathematical reasoning in open language models},
    author={Shao, Zhihong and Wang, Peiyi and Zhu, Qihao and Xu, Runxin and Song, Junxiao and Bi, Xiao and Zhang, Haowei and Zhang, Mingchuan and Li, YK and Wu, Yang and others},
    journal={arXiv preprint arXiv:2402.03300},
    year={2024}
}

@article{schulman2017proximal,
    title={Proximal policy optimization algorithms},
    author={Schulman, John and Wolski, Filip and Dhariwal, Prafulla and Radford, Alec and Klimov, Oleg},
    journal={arXiv preprint arXiv:1707.06347},
    year={2017}
}

@article{Yang2024Qwen25TR,
  title={Qwen2.5 Technical Report},
  author={Qwen An Yang and Baosong Yang and Beichen Zhang and Binyuan Hui and Bo Zheng and Bowen Yu and Chengyuan Li and Dayiheng Liu and Fei Huang and Guanting Dong and Haoran Wei and Huan Lin and Jian Yang and Jianhong Tu and Jianwei Zhang and Jianxin Yang and Jiaxin Yang and Jingren Zhou and Junyang Lin and Kai Dang and Keming Lu and Keqin Bao and Kexin Yang and Le Yu and Mei Li and Mingfeng Xue and Pei Zhang and Qin Zhu and Rui Men and Runji Lin and Tianhao Li and Tingyu Xia and Xingzhang Ren and Xuancheng Ren and Yang Fan and Yang Su and Yi-Chao Zhang and Yunyang Wan and Yuqi Liu and Zeyu Cui and Zhenru Zhang and Zihan Qiu and Shanghaoran Quan and Zekun Wang},
  journal={ArXiv},
  year={2024},
  volume={abs/2412.15115},
  url={https://api.semanticscholar.org/CorpusID:274859421}
}

@inproceedings{hu2022lora,
    title={Lo{RA}: Low-Rank Adaptation of Large Language Models},
    author={Edward J Hu and yelong shen and Phillip Wallis and Zeyuan Allen-Zhu and Yuanzhi Li and Shean Wang and Lu Wang and Weizhu Chen},
    booktitle={International Conference on Learning Representations},
    year={2022},
    url={https://openreview.net/forum?id=nZeVKeeFYf9}
}

@inproceedings{wolf2020transformers,
  title={Transformers: State-of-the-art natural language processing},
  author={Wolf, Thomas and Debut, Lysandre and Sanh, Victor and Chaumond, Julien and Delangue, Clement and Moi, Anthony and Cistac, Pierric and Rault, Tim and Louf, R{\'e}mi and Funtowicz, Morgan and others},
  booktitle={Proceedings of the 2020 conference on empirical methods in natural language processing: system demonstrations},
  pages={38--45},
  year={2020}
}

@inproceedings{kwon2023efficient,
  title={Efficient memory management for large language model serving with pagedattention},
  author={Kwon, Woosuk and Li, Zhuohan and Zhuang, Siyuan and Sheng, Ying and Zheng, Lianmin and Yu, Cody Hao and Gonzalez, Joseph and Zhang, Hao and Stoica, Ion},
  booktitle={Proceedings of the 29th symposium on operating systems principles},
  pages={611--626},
  year={2023}
}

@article{wang2026beyond,
  title={Beyond the 80/20 rule: High-entropy minority tokens drive effective reinforcement learning for llm reasoning},
  author={Wang, Shenzhi and Yu, Le and Gao, Chang and Zheng, Chujie and Liu, Shixuan and Lu, Rui and Dang, Kai and Chen, Xiong-Hui and Yang, Jianxin and Zhang, Zhenru and others},
  journal={Advances in Neural Information Processing Systems},
  volume={38},
  pages={115452--115486},
  year={2026}
}

@article{hendrycks2021measuring,
  title={Measuring mathematical problem solving with the math dataset},
  author={Hendrycks, Dan and Burns, Collin and Kadavath, Saurav and Arora, Akul and Basart, Steven and Tang, Eric and Song, Dawn and Steinhardt, Jacob},
  journal={arXiv preprint arXiv:2103.03874},
  year={2021}
}

@article{cobbe2021training,
  title={Training verifiers to solve math word problems},
  author={Cobbe, Karl and Kosaraju, Vineet and Bavarian, Mohammad and Chen, Mark and Jun, Heewoo and Kaiser, Lukasz and Plappert, Matthias and Tworek, Jerry and Hilton, Jacob and Nakano, Reiichiro and others},
  journal={arXiv preprint arXiv:2110.14168},
  year={2021}
}

@article{lewkowycz2022solving,
  title={Solving quantitative reasoning problems with language models},
  author={Lewkowycz, Aitor and Andreassen, Anders and Dohan, David and Dyer, Ethan and Michalewski, Henryk and Ramasesh, Vinay and Slone, Ambrose and Anil, Cem and Schlag, Imanol and Gutman-Solo, Theo and others},
  journal={Advances in neural information processing systems},
  volume={35},
  pages={3843--3857},
  year={2022}
}

@inproceedings{he2024olympiadbench,
  title={Olympiadbench: A challenging benchmark for promoting agi with olympiad-level bilingual multimodal scientific problems},
  author={He, Chaoqun and Luo, Renjie and Bai, Yuzhuo and Hu, Shengding and Thai, Zhen and Shen, Junhao and Hu, Jinyi and Han, Xu and Huang, Yujie and Zhang, Yuxiang and others},
  booktitle={Proceedings of the 62nd Annual Meeting of the Association for Computational Linguistics (Volume 1: Long Papers)},
  pages={3828--3850},
  year={2024}
}

@article{hendrycks2020measuring,
  title={Measuring massive multitask language understanding},
  author={Hendrycks, Dan and Burns, Collin and Basart, Steven and Zou, Andy and Mazeika, Mantas and Song, Dawn and Steinhardt, Jacob},
  journal={arXiv preprint arXiv:2009.03300},
  year={2020}
}

@article{wang2024mmlu,
  title={Mmlu-pro: A more robust and challenging multi-task language understanding benchmark},
  author={Wang, Yubo and Ma, Xueguang and Zhang, Ge and Ni, Yuansheng and Chandra, Abhranil and Guo, Shiguang and Ren, Weiming and Arulraj, Aaran and He, Xuan and Jiang, Ziyan and others},
  journal={Advances in Neural Information Processing Systems},
  volume={37},
  pages={95266--95290},
  year={2024}
}

@inproceedings{gema2025we,
  title={Are we done with mmlu?},
  author={Gema, Aryo Pradipta and Leang, Joshua Ong Jun and Hong, Giwon and Devoto, Alessio and Mancino, Alberto Carlo Maria and Saxena, Rohit and He, Xuanli and Zhao, Yu and Du, Xiaotang and Madani, Mohammad Reza Ghasemi and others},
  booktitle={Proceedings of the 2025 Conference of the Nations of the Americas Chapter of the Association for Computational Linguistics: Human Language Technologies (Volume 1: Long Papers)},
  pages={5069--5096},
  year={2025}
}

@article{du2026supergpqa,
  title={Supergpqa: Scaling llm evaluation across 285 graduate disciplines},
  author={Du, Xeron and Yao, Yifan and Ma, Kaijing and Wang, Bingli and Zheng, Tianyu and Liu, Minghao and Liang, Yiming and Jin, Xiaolong and Wei, Zhenlin and Zheng, Chujie and others},
  journal={Advances in Neural Information Processing Systems},
  volume={38},
  year={2026}
}

@article{rein2023gpqa,
  title={Gpqa: A graduate-level google-proof q\&a benchmark},
  author={Rein, David and Hou, Betty Li and Stickland, Asa Cooper and Petty, Jackson and Pang, Richard Yuanzhe and Dirani, Julien and Michael, Julian and Bowman, Samuel R},
  journal={arXiv preprint arXiv:2311.12022},
  year={2023}
}

\appendix

\section{Datasets}
\label{sec:datasets}
Our training data is designed to cover both general-domain and mathematical reasoning.

For general-domain reasoning, we use NEMOTRON-CROSSTHINK-QA from NEMOTRON-CROSSTHINK~\citep{akter-etal-2026-nemotron}. NEMOTRON-CROSSTHINK integrates multi-domain and multi-format corpora to support reasoning training beyond purely mathematical reasoning tasks. Rather than adopting the full corpus configuration, we focus on the synthetic datasets introduced in the original work and select NEMOTRON-CROSSTHINK-QA based on our data quality analysis.

For mathematical reasoning, we use MATH~\citep{hendrycks2021measuring}. This combination preserves the cross-domain design motivation of NEMOTRON-CROSSTHINK while covering both general-domain and mathematical reasoning, as well as both multiple-choice and open-ended question formats. The resulting dataset contains 21,204 training instances and 1,000 validation instances.

\paragraph{Scope of data analysis.}
This work focuses on the training method rather than on identifying an optimal mixture of training domains. We therefore keep the training data fixed across all RL-based methods and use the data mixture only as a controlled cross-domain training environment. A systematic study of domain composition and data ratio is orthogonal to our contribution and is left for future work.

\section{Benchmarks}
\label{sec:benchmarks}
For mathematical and STEM reasoning, we evaluate on GSM8K~\citep{cobbe2021training}, MATH-500~\citep{hendrycks2021measuring}, Minerva~\citep{lewkowycz2022solving}, OlympiadBench~\citep{he2024olympiadbench}, AMC 2023, GPQA~\citep{rein2023gpqa}, and GPQA-Diamond. 

For general-domain reasoning, we evaluate on MMLU~\citep{hendrycks2020measuring}, MMLU-Pro~\citep{wang2024mmlu}, MMLU-Redux~\citep{gema2025we}, and SuperGPQA~\citep{du2026supergpqa}. 

The source repository for each benchmark is provided in Table~\ref{tab:benchmarks}.

\section{Baselines}
\label{sec:baselines}
\paragraph{Base} This baseline refers to an off-the-shelf model that has undergone pretraining, supervised fine-tuning, and RLHF, but has not been further optimized with RL for reasoning. It serves as the initialization model for all RL-based methods.

\paragraph{Binary-RL} This baseline applies GRPO reasoning training with a binary outcome reward. A response receives a positive reward if its final answer matches the ground-truth answer and zero otherwise. This setting represents the standard outcome-level RLVR paradigm with sparse final-answer supervision.

\paragraph{Likelihood-RL} This baseline replaces the binary reward with an answer-conditioned likelihood reward. Specifically, each sampled reasoning response is scored using the likelihood of the ground-truth answer conditioned on the prompt and the generated reasoning prefix. Similar likelihood-based reward formulations have been explored in prior work~\citep{zhou2026reinforcing, yu2025rlpr, kwiatkowski2026likelihood}.

\paragraph{Scope of comparison.}
Our baselines are designed for controlled objective-level comparisons rather than full system-level reproductions of prior RLVR pipelines. Although prior methods could, in principle, be re-trained with the same data, backbone, and evaluation protocol, full RLVR systems typically involve method-specific components, such as verifier construction, reward shaping, sampling strategies, filtering rules, prompt formats, rollout budgets, and hyperparameter schedules. Adapting and tuning these components under a new data distribution would introduce additional design choices beyond the proposed mechanism studied in this work. We therefore instantiate representative objective-level baselines under a matched setting: Binary-RL captures sparse outcome-level RLVR, and Likelihood-RL captures answer-conditioned reward learning. This design isolates the additional contribution of token-level advantage modulation, while full system-level comparisons remain complementary.

\section{Detailed Results}
\label{sec:datailed_results}
This section provides the detailed benchmark results referenced in the main text. Table~\ref{tab:size-performance} reports the performance of GeneralThinker across different model scales, and Table~\ref{tab:detail-ablation-performance} presents the detailed ablation results.

\section{Hyperparameter Settings}
\label{sec:hyperparameter}
The hyperparameters used in this study are summarized in Table~\ref{tab:hyperparameter}.
\paragraph{Selection of $\epsilon_{\mathrm{tok}}$.}
The token-level clipping threshold $\epsilon_{\mathrm{tok}}$ is selected based on training stability rather than final benchmark accuracy. This is because clipping is introduced to suppress rare high-magnitude terms that may destabilize optimization. Moreover, the average clipping ratio remains below $5\%$, indicating that clipping only affects a small fraction of tokens rather than acting as a strong regularizer over the entire response.
\paragraph{Sensitivity to $\lambda_{\mathrm{tok}}$.}
We further examine the sensitivity of GeneralThinker to the token-level modulation strength. Since $\lambda_{\mathrm{tok}}$ directly controls the contribution of token-wise modulation to the final advantage, we vary it while keeping all other hyperparameters fixed. As shown in Table~\ref{tab:lambda_analysis}, GeneralThinker achieves the best average performance when $\lambda_{\mathrm{tok}}=0.1$, which suggests that a moderate degree of token-level modulation is beneficial for both mathematical \& STEM and general-domain reasoning. When $\lambda_{\mathrm{tok}}$ is too small, the token-level signal has limited influence, reducing the benefit of fine-grained credit assignment. In contrast, larger values such as 0.2 and 0.4 do not further improve performance and may introduce overly strong token-level perturbations to the response-level advantage. Overall, these results indicate that controlled token-level modulation is important: it should be strong enough to provide useful fine-grained guidance, but not so strong that it overwhelms the response-level learning signal.

\section{Computational Overhead}
\label{sec:overhead}

GeneralThinker introduces additional computation only in the scoring stage. Compared with standard GRPO with binary rewards, GeneralThinker requires two extra forward passes for each sampled response: one for computing the answer-conditioned likelihood reward $r_j^{\mathrm{LR}}$, and one for computing the answer-conditioned token-level modulation term $r^{\mathrm{tok}}_{j,t}$. Compared with Likelihood-RL, which already computes $r_j^{\mathrm{LR}}$, GeneralThinker adds only the second forward pass for token-level modulation.

Importantly, these additional scoring passes do not enlarge the optimization graph. All rewards, reward-derived advantages, and token-level modulation terms are detached before the update. Therefore, the backward pass is still performed only through the current-policy likelihood ratios and the KL regularization term. In practice, this means that GeneralThinker increases scoring-time forward computation, but does not introduce additional reward-gradient paths or extra backward computation through the reward model.

GeneralThinker also reuses quantities already required by GRPO optimization when possible. In particular, the reference-policy log probabilities used in the token-level compatibility score are the same type of quantities used for KL regularization. Thus, the main additional computation beyond Likelihood-RL is the answer-conditioned policy forward pass for token-level scoring. This design keeps the response-level likelihood advantage as the primary learning signal while adding fine-grained credit assignment with limited additional optimization cost.

\onecolumn

\begin{table}
    \centering
    \begin{tabular}{cll}
        \Xhline{1.0pt}
        \textbf{Domain} & \textbf{Name} & \textbf{Repository (https://huggingface.co/datasets/)}\\ 
        \Xhline{0.5pt}

        \multirow{7}{*}{Math \& STEM} 
            & GSM8K         & openai/gsm8k \\
            & MATH-500      & nlile/hendrycks-MATH-benchmark \\
            & Minerva       & math-ai/minervamath \\
            & OlympiadBench & math-ai/olympiadbench \\
            & GPQA          & Idavidrein/gpqa \\
            & GPQA-Diamond  & Idavidrein/gpqa \\
            & AMC23         & math-ai/amc23 \\
        \Xhline{0.5pt}
        \multirow{4}{*}{General}
            & MMLU           & cais/mmlu \\
            & MMLU-Pro       & TIGER-Lab/MMLU-Pro \\
            & MMLU-Redux     & edinburgh-dawg/mmlu-redux-2.0 \\
            & SuperGPQA & m-a-p/SuperGPQA \\

        \Xhline{1.0pt}
    \end{tabular}
    \caption{Summary of evaluation benchmarks by domain, including dataset names and Hugging Face repositories.}
    \label{tab:benchmarks}
\end{table}

\begin{table}
    \centering
    \begin{tabular}{lccc}
        \Xhline{1.0pt}
        & \textbf{GeneralThinker-1.5B} & \textbf{GeneralThinker-3B} & \textbf{GeneralThinker-7B}\\ \Xhline{0.5pt}
        \multicolumn{4}{c}{\textbf{Math \& STEM}} \\
        MATH-500        & 54.00             & 59.40             & 73.60 \\ 
        GSM8K           & 72.40             & 84.61             & 91.96 \\ 
        Minerva         & 20.59             & 26.84             & 35.29 \\
        OlympiadBench   & 21.07             & 26.56             & 36.50 \\
        AMC23           & 30.00             & 25.00             & 45.00 \\
        GPQA            & 20.54             & 18.75             & 25.00 \\
        GPQA-Diamond    & 19.19             & 28.28             & 26.77 \\
        \cdashline{1-4}
        \textit{Average}    & 33.97            & 38.49             & 47.73 \\
        \Xhline{0.5pt}
        \multicolumn{4}{c}{\textbf{General}} \\
        MMLU            & 53.50             & 62.78             & 72.71 \\
        MMLU-Pro        & 27.13             & 36.36             & 45.80 \\
        MMLU-Redux      & 54.72             & 63.54             & 73.81 \\
        SuperGPQA       & 12.79             & 16.76             & 20.76 \\
        \cdashline{1-4}
        \textit{Average}    & 37.04            & 44.86             & 53.27 \\
        \Xhline{1.0pt}
    \end{tabular}
    \caption{Accuracy (\%) of GeneralThinker models of different sizes on Math \& STEM and General benchmarks.}
    \label{tab:size-performance}
\end{table}

\begin{table}
\renewcommand{\arraystretch}{1}
    \centering
    \begin{tabular}{lccccc}
        \Xhline{1.0pt}
        & \textbf{Likelihood-RL} & \textbf{AM only} & \textbf{AM+TC} & \textbf{AM+DP} & \textbf{Full}\\
        \Xhline{0.5pt}
        \multicolumn{6}{c}{\textbf{Math \& STEM}} \\
        MATH-500        & 54.80        & 54.60        & 53.00        & 53.20        & 54.00\\ 
        GSM8K           & 72.33        & 71.57        & 71.79        & 72.25        & 72.40\\ 
        Minerva         & 19.85        & 21.32        & 20.59        & 20.96        & 20.59\\
        OlympiadBench    & 19.88        & 19.73        & 21.22        & 20.62        & 21.07\\
        AMC23           & 20.00        & 22.50        & 30.00        & 25.00        & 30.00\\
        GPQA            & 15.40        & 17.63        & 18.08        & 16.74        & 20.54\\
        GPQA-Diamond    & 18.18        & 16.16        & 16.16        & 17.17        & 19.19\\
        \cdashline{1-6}
        \textit{Average}    & 31.49    & 31.93        & 32.98        & 32.28        & 33.97\\
        \Xhline{0.5pt}
        \multicolumn{6}{c}{\textbf{General}} \\
        MMLU            & 52.61        & 49.36        & 49.69        & 50.16        & 53.50\\
        MMLU-Pro        & 26.32        & 24.62        & 24.78        & 25.29        & 27.13\\
        MMLU-Redux      & 53.77        & 50.89        & 51.05        & 52.09        & 54.72\\
        SuperGPQA       & 12.76        & 12.35        & 12.61        & 12.74        & 12.79\\
        \cdashline{1-6}
        \textit{Average}    & 36.37    & 34.31        & 34.53        & 35.07        & 37.04\\
        \Xhline{1.0pt}
    \end{tabular}
    \caption{Ablation results of GeneralThinker.}
    \label{tab:detail-ablation-performance}
\end{table}

\begin{table}
    \centering
    \begin{tabular}{lcccc}
        \Xhline{1.0pt}
        & \textbf{0.05} & \textbf{0.1} & \textbf{0.2} & \textbf{0.4}\\ \Xhline{0.5pt}
        \multicolumn{5}{c}{\textbf{Math \& STEM}} \\
        MATH-500        & 53.20             & 54.00             & 53.20             & 54.00 \\ 
        GSM8K           & 71.80             & 72.40             & 71.87             & 71.72\\ 
        Minerva         & 20.22             & 20.59             & 18.75             & 19.85\\
        OlympiadBench   & 19.58             & 21.07             & 21.36             & 21.07\\
        AMC23           & 22.50             & 30.00             & 25.00             & 27.50\\
        GPQA            & 15.40             & 20.54             & 15.40             & 17.19\\
        GPQA-Diamond    & 17.17             & 19.19             & 20.20             & 15.66\\
        \cdashline{1-5}
        \textit{Average}    & 31.41            & 33.97             & 32.25             & 32.43\\
        \Xhline{0.5pt}
        \multicolumn{5}{c}{\textbf{General}} \\
        MMLU            & 49.27             & 53.50             & 49.67             & 48.77\\
        MMLU-Pro        & 25.13             & 27.13             & 25.02             & 23.79\\
        MMLU-Redux      & 51.25             & 54.72             & 51.72             & 50.65\\
        SuperGPQA       & 12.40             & 12.79             & 12.36             & 12.62\\
        \cdashline{1-5}
        \textit{Average}    & 34.51            & 37.04             & 34.96             & 33.96\\
        \Xhline{1.0pt}
    \end{tabular}
    \caption{Accuracy (\%) of GeneralThinker with different modulation strengths.}
    \label{tab:lambda_analysis}
\end{table}

\begin{table}
    \centering
    \begin{tabular}{ccl}
        \Xhline{1.0pt}
        \textbf{Hyperparameter} & \textbf{Value} & \textbf{Note}\\ \Xhline{0.5pt}
        \multicolumn{3}{c}{\textbf{\textit{Training Configuration}}} \\
        epochs & 1 & Number of training epochs. \\
        batch\_size & 64 & Batch size. \\
        lr & $1\times10^{-6}$ & Learning rate for the AdamW optimizer. \\
        \multicolumn{3}{c}{\textbf{\textit{GRPO Configuration}}} \\
        $\beta$ & 0.001 & The weight for KL regularization. \\
        $G$ & 8 & Group size used for response generation. \\
        temperature & 1.0 & The value used to modulate the next-token probabilities. \\
        max\_length & 1024 & Maximum generation length. \\
        \multicolumn{3}{c}{\textbf{\textit{GeneralThinker Configuration}}} \\
        $\lambda_{\mathrm{tok}}$ & 0.1 & The strength of token-level modulation. \\
        $\epsilon_{\mathrm{tok}}$ & 2.0 & The threshold of token clipping. \\
        \multicolumn{3}{c}{\textbf{\textit{LoRA Configuration}}} \\
        $r$ & 32 & Rank. \\
        $\mathrm{lora\_}\alpha$ & 64 & Scaling factor. \\
        \Xhline{1.0pt}
    \end{tabular}
    \caption{Summary of hyperparameter settings.}
    \label{tab:hyperparameter}
\end{table}

\begin{table}[t]
\centering
\begin{tcolorbox}[
    colback=gray!5!white,
    colframe=black,
    title=\textbf{\textsc{Input Template of Likelihood Reward}},
    width=0.95\linewidth,
    boxrule=1pt
]
\{Question\} \newline
Please reason step by step, and put your final answer within \textbackslash boxed\{\}.
\end{tcolorbox}
\caption{Prompt template used for response generation before likelihood-reward scoring.}
\label{tab:likelihood_prompt}
\end{table}

\begin{table}[t]
\centering
\begin{tcolorbox}[
    colback=gray!5!white,
    colframe=black,
    title=\textbf{\textsc{Input Template of Token-Level Advantage Modulation}},
    width=0.95\linewidth,
    boxrule=1pt
]
\{Question\} \newline
Please reason step by step given the final answer within \textbackslash boxed\{\}. \newline
\textbackslash boxed\{Ground-truth answer\}
\end{tcolorbox}
\caption{Prompt template used for token-level advantage modulation.}
\label{tab:token_prompt}
\end{table}

\end{document}